\documentclass[runningheads]{llncs}
\usepackage{graphicx}
\usepackage{caption}
\begin{document}
\title{Automatic Cobb Angle Detection using Vertebra Detector and Vertebra Corners Regression}
%
%
\author{Bidur Khanal\inst{1}\and
Lavsen Dahal\inst{1}\and
Prashant Adhikari\inst{2}\and
Bishesh Khanal\inst{1}}
\titlerunning{Automatic Cobb Angle Detection}
\authorrunning{B. Khanal \textit{et al.}}
%

\institute{NepAl Applied Mathematics and Informatics Institute for Research (NAAMII) \\
\email{\{bidur.khanal, lavsen.dahal, bishesh.khanal\}@naamii.org.np} \and
Hospital for Advanced Medical Surgery (HAMS), Kathmandu, Nepal\\}

\maketitle              

\newcommand{\LD}[1]{\textcolor{red}{#1}}

\begin{abstract}
Correct evaluation and treatment of Scoliosis require accurate estimation of spinal curvature. Current gold standard is to manually estimate Cobb Angles in spinal X-ray images which is time consuming and has high inter-rater variability. We propose an automatic method with a novel framework that first detects vertebrae as objects followed by a landmark detector that estimates the 4 landmark corners of each vertebra separately. Cobb Angles are calculated using the slope of each vertebra obtained from the predicted landmarks. For inference on test data, we perform pre and post processings that include cropping, outlier rejection and smoothing of the predicted landmarks. The results were assessed in AASCE MICCAI challenge 2019 which showed a promise with a SMAPE score of 25.69 on the challenge test set.

\keywords{Scoliosis  \and Landmark \and Object Detection \and Cobb Angle.}
\end{abstract}
\bibliographystyle{unsrt}   

\section{Introduction}
{\label{sec:intro}}
Scoliosis is a sideways curvature of the spine occurring mostly in teens. 
Severe scoliosis can also lead to disability.
The current gold standard for diagnosing scoliosis is manual measurement of Cobb Angles in anterior-posterior (AP) or lateral (LAT) X-ray images which involve identifying the most tilted vertebrae above and below the apex of the spinal curve \cite{Greiner2002}.
However, the procedure is time-consuming and observer dependent, leading to high inter-observer variability that could negatively impact assessing prognosis and treatment decisions \cite{Loder_2004}.
Thus, there has been increasing interest in automatic estimation of Cobb angles directly from the X-ray images.
In this context, we participated in MICCAI 2019 challenge on Accurate Automated Spinal Curvature Estimation (AASCE) \footnote{https://aasce19.grand-challenge.org/Home/} where the task was to accurately estimate three Cobb angles \cite{Sardjono2013} from the training dataset containing 609 AP x-rays \footnote{\label{spineweb}http://spineweb.digitalimaginggroup.ca/spineweb/index.php?n=Main.Datasets} whose results were assessed on 98 test images.
The ground truth (GT) annotations are the anatomical landmarks consisting of four corners of 17 vertebrae: twelve thoracic and five lumbar.

\paragraph{Related Work:} The two most common approaches of estimating Cobb angles are Segmentation based and Landmark based approaches.
The segmentation based methods first segment all the vertebrae or the end-plates of the vertebrae to identify the most tilted vertebrae from which the Cobb angles are estimated \cite{Sardjono2013,Allen2008,Zhang2009}.
Accurate segmentation of each vertebra from X-ray images is difficult with traditional feature-engineering based approaches. To our knowledge, even modern supervised deep neural networks are not robust and accurate enough yet for the vertebra segmentation. Creating accurate GT segmentation is time consuming and relatively difficult compared to annotating landmarks:~four corners of the vertebrae.
In Landmark based approach which is the state-of-the-art, the four corners of each vertebrae are detected and are subsequently used for estimating Cobb angles.
Some methods jointly estimate all the landmarks and Cobb angles, while others first estimate landmarks followed by Cobb angle computation which might include outlier rejection and post-processing techniques \cite{Wu_2017,Sun_2017}.

There are several approaches of detecting landmarks in medical images such as Reinforcement learning \cite{Alansary_2019}, iterative patch based approaches \cite{Li_2018} and fully convolutional neural network based approaches~\cite{payer2019integrating}.
One important difference in vertebra landmarks compared to other anatomical landmarks is the presence of a large number of similar looking vertebrae.
We believe that detecting vertebrae as objects before finding landmarks within the detected vertebrae is advantageous as it allows:
i) avoiding difficulty for translation equivariant CNNs to learn very different coordinate locations for almost identical appearing vertebrae
ii) leveraging popular object detectors
pre-trained for computer vision tasks
iii) Reducing the search space for landmark detector

\paragraph{Contribution:} We propose a novel approach to first detect 17 vertebrae with a bounding box object detector, after which each of the predicted boxes is fed to a landmark detector as illustrated in Figure {\ref{fig:pipeline}}. The predicted landmarks are post-processed to remove outliers before calculating the three Cobb angles.
\cite{Ruhan2017} used Faster-RCNN \cite{Ren_2017} object detector to detect intervertebral disc in lateral X-rays, but they left the landmark detection as a future work.

\section{Dataset}
{\label{dataset}}

The dataset consists of 609 spinal AP x-ray images available at SpineWeb \footnote{\label{spineweb1}http://spineweb.digitalimaginggroup.ca/spineweb/index.php?n=Main.Datasets} as \textit{Dataset 16}. Each image has 68 GT landmarks corresponding to 4 corners of the 17 vertebrae, and 3 Cobb Angles.
Organizers provided test images without GT separately.
We connected the four landmark corners of each vertebrae to create a box whose width and height were then increased symmetrically by 50 and 10 pixels respectively to create GT bounding boxes.
All the bounding boxes were labelled as belonging to a single class.
The GT bounding boxes were used to crop and extract individual vertebrae as a single separate image containing four landmark corners.
The coordinates of the landmarks are normalized to the coordinate system that maps all the pixel coordinates of the cropped image to the interval {[}0, 1{]}.
The normalized landmark coordinates are used as GT labels for the landmark regression network.

\section{Vertebrae Detection followed by Landmarks Regression}
\begin{figure}
\includegraphics[width=\textwidth]{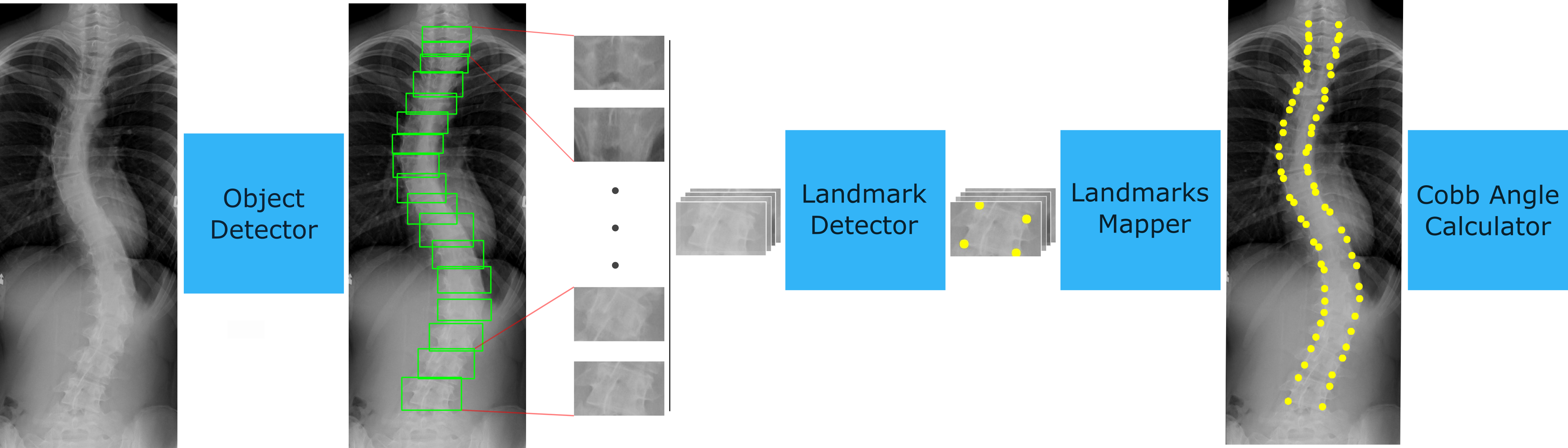}
\caption{Proposed Framework: The input images are first passed to an object detector that detects the vertebrae. The detected vertebrae are extracted as individual images and passed to a landmark detector that detects the four corners of the vertebra. The landmarks are mapped back to the original image from which Cobb angles are calculated. We use CNN-based Faster-RCNN and DenseNet for object detection and landmark detection respectively.}
\label{fig:pipeline}
\vspace{-4mm}
\end{figure}

{\label{sec:overall_pipeline}}
We use an object detector to detect the vertebrae as bounding box objects which are then fed to a landmark regression network as separate input images. The predicted normalized landmark coordinates from individual bounding boxes are combined and mapped back to the original images as shown in Figure \ref{fig:pipeline}.

\subsection{Training Vertebra Detection with Faster-RCNN }
{\label{subsec:FRCNN}}
Faster-RCNN \cite{Ren_2017} is a widely used two-stage object detector consisting of: i) a Region Proposal Network (RPN) that proposes potential object regions from a set of anchor boxes of various sizes in a sliding window over the feature maps extracted from a CNN-based base network ii) a fully connected and a bounding box regression layer that regress bounding box locations of the identified objects. 
We used ResNet V1 101 with pre-trained weights on Imagenet data \footnote{https://github.com/tensorflow/models/tree/master/research/slim} as the base network, which was fine-tuned after block 2.
We used two scales with box areas of 64\textsuperscript{2} and 128\textsuperscript{2} pixels, and aspect ratios 1:1 and 2:1 for RPN's anchor boxes, as the vertebrae are relatively small and do not have extreme aspect ratios.
The network was trained for around 180k steps with batch size 1 using SGD optimizer with momentum 0.9, learning rate 0.0003 and early stopping.
The implementation was adopted from Luminoth\footnote{https://github.com/tryolabs/luminoth} in Tensorflow  framework 10.1.
Data augmentation included random Gaussian noise ($\mu=0$, $\sigma=0.005$), and vertical and horizontal flips with a probability of 0.5.
All the images were rescaled preserving the aspect ratio such that its sizes remained within 600 - 1000 pixels as much as possible.

\subsection{Training Landmark Detector with DenseNet}
{\label{Densenet}}
The four corner landmarks were estimated using a Densely Connected Convolutional Neural Network (DenseNet) which are
known to require fewer parameters than traditional CNN \cite{Huang_2017}.
In DenseNet, each layer's feature maps are used for all subsequent layers within a block, where each block constitutes a bottleneck layer (a 2d Convolution layer with 1x1 filter size), batch normalization, ReLU activation, and a regular 2D convolution layer (3x3 filter size).
We used 5 blocks with a growth rate of 8 which is the number of output feature maps of each layer.
The 2D Global Average Pooling is used after 5 blocks followed by a dense layer.
The final layer consists of 8 output units with a linear activation function.
All the input images to landmark detector were resized to 200 x 120 pixels.

\section{Pre and Post Processing During Inference}
{\label{sec:processing}}
\paragraph{Cropping:}
Almost all test images contained skull and pelvic regions but none of the training images had them.
During training, the model did not see negative samples of skull and pelvic regions making it prone to falsely detect structures appearing similar to vertebra such as jaws.
We randomly picked one test image with an aspect ratio $a_0$ and found empirically that cropping ${c_{t_0}} = 0.18$ and $ {c_{b_0}}=0.21$ times the image height from the top and bottom removed skull and pelvic regions satisfactorily.
All the remaining test images with aspect ratio $a$ were cropped by $ {c_{t}}$= $ {c_{t_0}}\cdot \frac{a}{a_{0}}$ and $ {c_{b}} $= $ {c_{b_0}}\cdot \frac{a}{a_{0}}$ fraction of the image height from the top and bottom respectively.
\paragraph{Outlier Rejection:}
We removed some of the outliers by using the fact that adjacent vertebrae cannot be far away from each other: if the x-center (horizontal) of any detected
bounding box is more than half box width away from the x-centers of both of its two nearest neighboring (top and bottom) boxes, they are rejected as outliers.
For the topmost and bottom boxes, the same test was done against only one nearest
neighbor.

\paragraph{Curve fitting and Cobb Angle Calculation from Predicted Landmarks:}
We used the code provided along with the challenge dataset \cite{Wu_2017} to calculate 3 Cobb angles - Main Thoracic (MT), Proximal Thoracic (PT) and Thoracolumbar/Lumbar (TL/L) from a given set of landmarks.
It did not work well when the number of landmarks were not exactly 68 corresponding to the 17 bounding boxes.
To ensure exactly 68 landmark points for angle calculation, we used the following after outlier rejection:
when the detected vertebrae number is more than 17, reject extra bounding boxes starting from the bottom.
Similarly, if the number is less than 17, duplicate the bottom landmarks as required.
We also smoothed the landmarks by fitting a polynomial curve where the degree 6 polynomial gave best fit out of 3 to 8 on visual inspection.
The x-coordinate of each landmark is regressed by using the y-coordinate as the independent variable of the fitted polynomial.
The smoothed landmarks were the ones that were used to estimate Cobb angles in the final test score.

\section{Results}
{\label{results}}
The results were evaluated with symmetric mean
absolute percentage error, SMAPE$= \frac{1}{N}\sum_N{\frac{\sum_m{|{a_g - a_p}|}}{\sum_m{(a_g+a_p)}}} 100\%$, where we have $N(=98)$ test images, $m(=3)$ Cobb angles per image, GT angle $a_{g}$ and the corresponding predicted angle $a_{p}$.

\vspace*{-3mm}
\begin{table}
\caption{Results for different experiment setups}
\centering
\vspace{-2mm}
\begin{tabular}{|l|p{9.5cm}|l|l} 
\cline{1-3}
Exp no. & Processing for Test Images                                            & SMAPE   &   \\ 
\cline{1-3}
1       & No Cropping & 33.3\%  &   \\ 
\cline{1-3}
2       & Cropping and outlier removal without smoothing & 26.79\% &   \\ 
\cline{1-3}
3       & Cropping, outlier removal and smoothing with order 6 polynomial fitting & 25.69\% &   \\
\cline{1-3}
\end{tabular}
\label{table:result}
\vspace{-5mm}
\end{table}

\begin{figure}
\includegraphics[width=\textwidth]{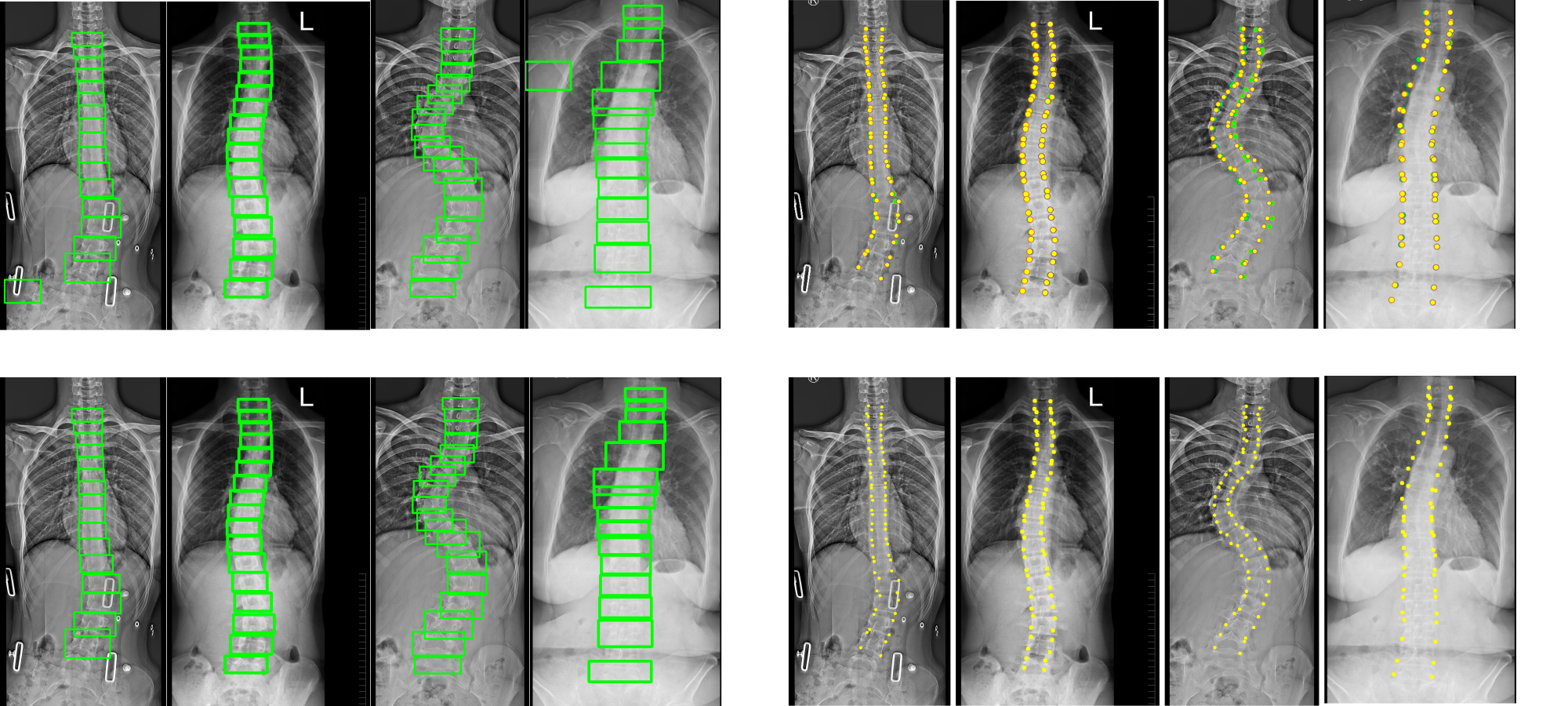}
\caption{Anti-clockwise from top left: bounding box detection, outlier rejection, landmark prediction and smoothing of landmarks (green) with polynomial degree 6 for four images of the test set}
\label{fig:bbox_lm_outlier_smoothened}
\vspace{-4mm}
\end{figure}

Table \ref{table:result} shows the results of three different experiments where we achieved our best score in the challenge by cropping, rejecting outliers and smoothing the estimated landmarks.
The top score in the leader board was 21.71\% when the challenge results entry was closed. Figure \ref{fig:bbox_lm_outlier_smoothened} shows detected bounding boxes and landmarks, and results of outlier rejection and smoothing with polynomial fitting in 4 example images from test set.  

\section{Discussion and Conclusion}
{\label{sec:discussion}}
Detecting vertebrae as objects before predicting corner landmarks is found to be a promising approach.
However, cropping all test images will not generalize well.
A more robust object detector trained with images having negative samples from skull and pelvic regions could eliminate the need of cropping.
The proposed approach does not properly take into account the inter-dependency between landmark positions of different vertebrae.
A learning algorithm to learn this inter-dependency could improve the results.
Finally, learning to estimate the angles directly from landmarks instead of using the geometric algorithm could be robust to noisy landmark prediction.

\paragraph{\textbf{Acknowledgements}} This work is supported by NVIDIA GPU donation. We also thank Pro-Mech Minds \& Engineering Services for agreeing to partially fund conference visit expenses for presenting this work.

%

%
%
%

\bibliography{bibliography_file}






\end{document}